\def\year{2020}\relax
\documentclass[letterpaper]{article} 
\usepackage{aaai20}  
\usepackage{times}  
\usepackage{helvet} 
\usepackage{courier}  
\usepackage[hyphens]{url}  
\usepackage{graphicx} 
\usepackage{graphicx}  
\title{Boosting Throughput and Efficiency of Hardware Spiking Neural Accelerators using Time Compression Supporting Multiple Spike Codes}
\author{Changqing Xu\textsuperscript{1}, Wenrui Zhang\textsuperscript{2}, Yu Liu\textsuperscript{3}, Peng Li\textsuperscript{4\thanks{Corresponding Email: lip@ucsb.edu}}\\ 
\textsuperscript{1,2,4}Department of Electrical \& Computer Engineering, University of California, Santa Barbara\\
\textsuperscript{3}Department of Electrical \& Computer Engineering, Texas A\&M University\\
\textsuperscript{1}changqingxu1020@163.com, \textsuperscript{2}wenruizhang@ucsb.edu, \textsuperscript{4}lip@ucsb.edu\\
}

\begin{document}

\maketitle

\begin{abstract}
Spiking neural networks (SNNs) are the third generation of neural networks and can explore both rate and temporal coding for energy-efficient event-driven computation. However, the decision accuracy of existing SNN designs is contingent upon processing a large number of spikes over a long period. Nevertheless, the switching power of SNN hardware accelerators is proportional to the number of spikes processed while the length of spike trains limits throughput and static power efficiency. This paper presents the first study on developing temporal compression to significantly boost throughput and reduce energy dissipation of digital hardware SNN accelerators while being applicable to multiple spike codes. The proposed compression architectures consist of low-cost input spike compression units, novel input-and-output-weighted spiking neurons, and reconfigurable time constant scaling to support large and flexible time compression ratios. Our compression architectures can be transparently applied to any given pre-designed SNNs employing either rate or temporal codes while incurring minimal modification of the neural models, learning algorithms, and hardware design. Using spiking speech and image recognition datasets, we demonstrate the feasibility of supporting large time compression ratios of up to 16$\times$, delivering up to 15.93$\times$, 13.88$\times$, and 86.21$\times$ improvements in throughput, energy dissipation, the tradeoffs between hardware area, runtime, energy, and classification accuracy, respectively based on different spike codes on a Xilinx Zynq-7000 FPGA. These results are achieved while incurring little extra hardware overhead. 
\end{abstract}

\section{Introduction}
Spiking neural networks (SNNs) closely emulate the spiking behaviors of biological brains \cite{1}. Moreover, the event-driven nature of SNNs offer potentials in achieving great computational/energy efficiency on hardware neuromorphic computing systems \cite{2,3}. For instance, processing a single spike may only consume a few pJ of energy on recent neuromorphic chips such as IBM’s TrueNorth \cite{2} and Intel’s Loihi \cite{4}.

SNNs  support various rate/temporal spike codes among which rate coding using Poisson spike trains is popular. However, in that case, the low-power advantage of SNNs may be offset  by long latency during which many spikes are processed for ensuring decision  accuracy. Various temporal codes have been attempted to improve the efficiency of information representation \cite{5,6,7,8,9}. The time-to-first-spike coding encodes information using arrival time of the first spike \cite{5}. Phase coding \cite{6} encodes information in a spike by its phase relative to a periodic reference signal \cite{7}. 
No coding is considered universally optimal thus far. The achievable latency/spike reduction of a particular code can vary widely with network structure and application. 
\begin{figure}[htbp]
  \centering
   \includegraphics[width=\linewidth]{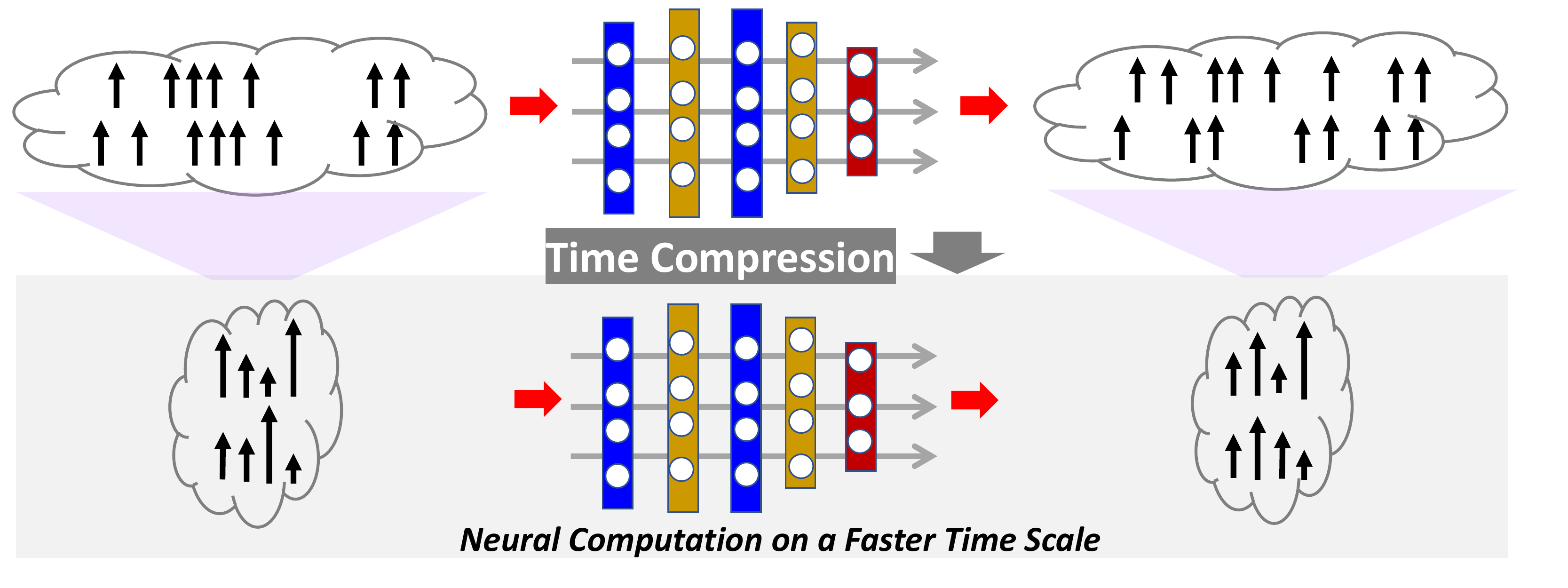}
  \caption{Proposed general time compression for SNNs.}
 \label{Fig_Compress}
\end{figure}

Rather than advocating a particular code, \emph{for the first time}, we focus on an orthogonal problem:  temporal compression applicable to any given SNN (accelerator) and spike code to boost throughput and  energy efficiency. We propose a general  compression technique  that preserves both the spike count and  temporal characteristics of the original SNN with low information loss, as shown in Fig.~\ref{Fig_Compress} It transparently compresses  duration of the spike trains, hence classification latency, on top of an existing rate/temporal code. More broadly, this work extends the notion of weight/model pruning/compression of DNN accelerators from the spatial domain to the temporal domain.

The  contributions of this paper include: \textbf{1)} the first general time-compression technique transparently compressing spike train duration of a given SNN and achieving large latency reduction on top of the spike codes that come with the SNN, \textbf{2)} facilitating the proposed time compression by four key ideas:  spike train compression using a weighted representation, a new family of input-output-weighted (IOW) spiking neural models for processing time-compressed spike trains for multiple spike codes,  scaling of time constants defining neural, synaptic, and learning dynamics, and low-cost support of flexible compression ratios (powers of two or not) using time averaging, \textbf{3)} low-overhead hardware modifications of a given SNN accelerator to operate it on a compressed time scale while preserving the spike counts and temporal behaviors in inference and training, \textbf{4)} a time-compressed SNN (TC-SNN) accelerator architecture and its programmable variant (PTC-SNN) operating on a wide range of (programmable) compression ratios and achieving significantly improved latency, energy efficiency,  and tradeoffs between latency/energy/classification accuracy.  

We demonstrate the proposed TC-SNN and PTC-SNN compression architectures by realizing several liquid-state machine (LSM) spiking neural accelerators with a time compression ratio up to 16:1 on a Xilinx Zynq-7000 FPGA. Using the TI46 Speech Corpus \cite{10}, the CityScape image recognition dataset \cite{11}, and N-TIDIGITS18 dataset \cite{anumula2018feature}, we demonstrate the feasibility of supporting large time compression ratios of up to 16$\times$, delivering up to 15.93$\times$, 13.88$\times$, and 86.21$\times$  improvements in throughput, energy dissipation, the tradeoffs between hardware area, runtime, energy, and classification accuracy, respectively  based  on various spike coding mechanisms including burst coding \cite{park2019fast} on a Xilinx Zynq-7000 FPGA. These results are achieved while incurring little extra hardware overhead.

\section{Proposed Time-Compressed Neural Computation}

This work aims to enable time-compressed neural computation that preserves the spike counts and temporal behaviors in inference and training of a given SNN while significantly improving latency, energy efficiency, and tradeoffs between latency/energy/classification accuracy.   \textbf{We develop four  techniques for  this objective}: 1) spike train compression using a weighted representation, 2) a new family of input-output-weighted (IOW) spiking neural models processing time-compressed spike trains for multiple spike codes, 3) scaling of time constants of neural, synaptic, and learning dynamics, and 4) low-cost support of flexible compression ratios (powers of two or not) using time averaging.



\subsection{Spike Train Compression in Weighted Form}
\begin{figure}[htbp]
  \centering
   \includegraphics[width=0.9\linewidth]{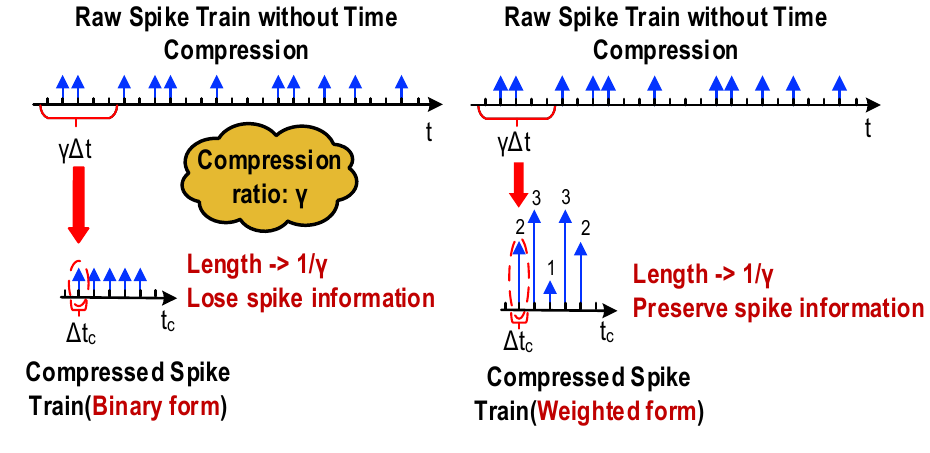}
  \caption{Binary vs.  (compressed) weighted spike trains.}
 \label{WeightInput}
\end{figure}

We time-compress a given spiking neural network first by  shrinking the duration of the input spike trains. To support large compression ratios hence significant latency reductions, we represent the compressed input trains using an weighted form. Typical binary spike trains with temporal sparsity  may be time-compressed into another binary spike train of a shorter duration. However, as shown in Fig.~\ref{WeightInput},  the spike count and temporal characteristics of the uncompressed train can only be preserved  under a small compression ratio bound by the minimal interspike interval.  More aggressive compression would lead to  merging multiple adjacent spikes into a single spike, resulting in significant alterations of firing count and temporally coded information. This severely limits the amount of compression possible. Instead, we propose a new weighted form for representing compressed spike trains, where multiple adjacent binary spikes are compressed into a single weighted spike with a weight value equal to the number of binary spikes combined, allowing preservation of spike information even under very large compression ratios (Fig.~\ref{WeightInput}). 

\subsection{Input-Output-Weighted (IOW) Spiking Neurons}
\begin{figure}[htbp]
  \centering
  \includegraphics[width=0.8\linewidth]{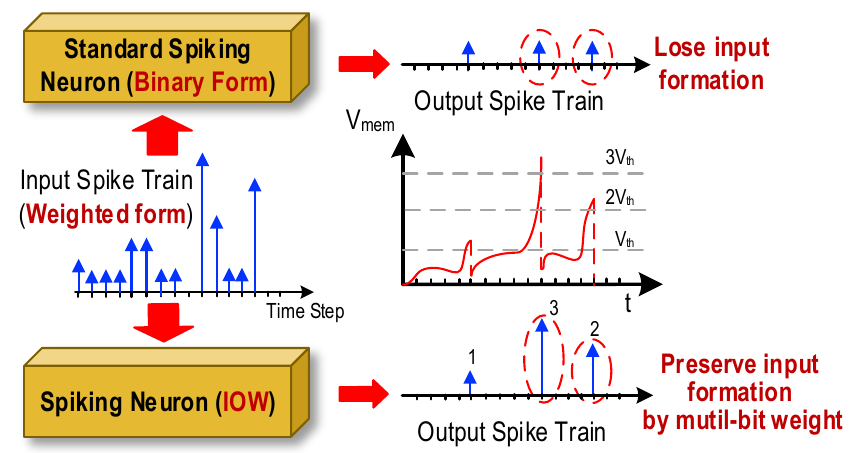}
  \caption{Binary vs. weighted output spikes. }
 \label{WeightOutput}
\end{figure}

As such, each spiking neuron would process the received input spike trains in the weighted form. Furthermore, as shown in Fig.~\ref{WeightOutput}, under large compression ratios the membrane potential of a spiking neuron may rise high above the firing threshold voltage  within a single time step as a result of receiving input spikes with large weights. In this case, outputting spike trains in the standard binary form can lead to significant loss of input formation, translating into large performance loss as we demonstrate in our experimental results. Instead, we propose a new  family of input-output-weighted (IOW) spiking neural models which take the input spike trains in the weighted form and produce the output spike train in the same weighted form, where the multi-bit weight value of each output spike reflects the amplitude of the membrane potential as a multiple of the firing threshold. Spiking neuronal models such as the leaky integrate-and-fire (LIF) model and other models  supporting various spike codes can be converted to their IOW counterpart with  streamlined low-overhead modification as detailed later.

\subsection{Scaling of Time Constants of SNN Dynamics}
The proposed compression is general in the sense that it intends to preserve the spike counts and temporal behaviors in the
neural dynamics, synaptic responses, and dynamics employed in the given SNN such that no substantial alterations are introduced by compression other than that the time-compressed SNN just effectively operates on a faster time scale. The dynamics of the cell membrane is typically specified by a membrane time constant $\tau_m$, which controls the process of action potential (spike) generation  and  influences the information processing of each spiking neuron \cite{12}. Synaptic models also play an important role in an SNN and may be specified by one or multiple time constants, translating received spike inputs into a continuous synaptic current waveform based on the dynamics of a particular order \cite{12}. Finally, Spike traces or temporal variables filtered with a specific time constant may be used to implement spike-dependent learning rules \cite{5,17}. 

\begin{figure}[htbp]
   \centering
   \includegraphics[width=\linewidth]{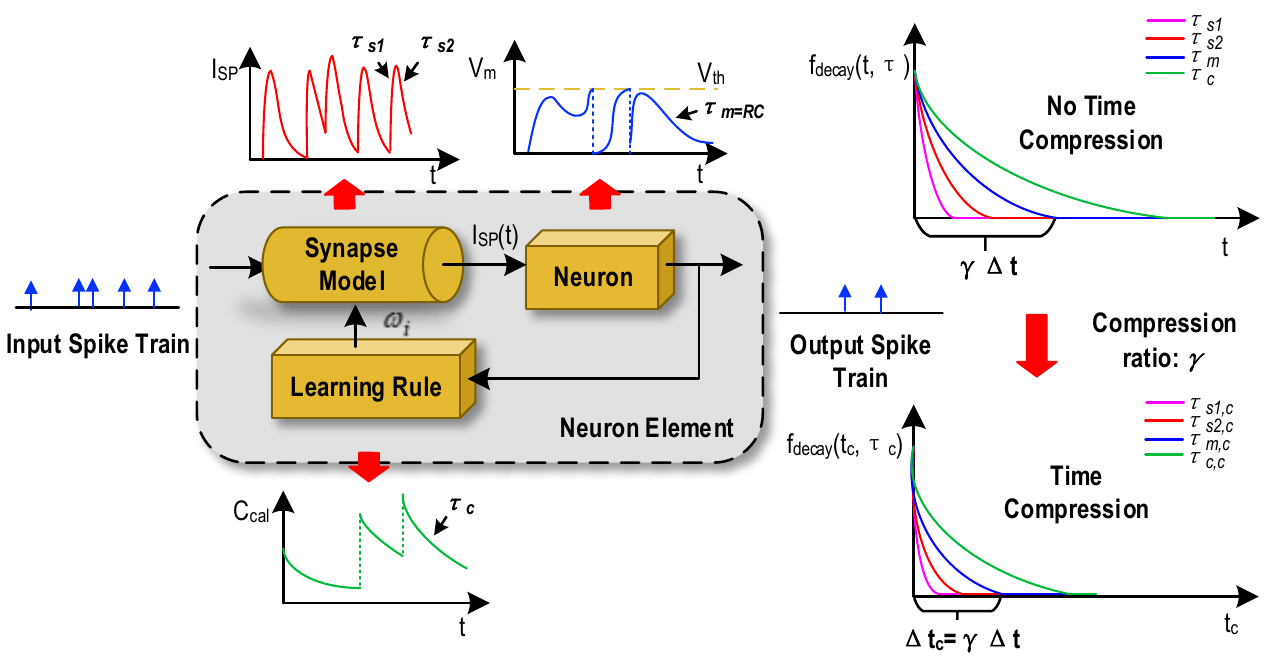}
   \caption{Scaling of time constants of SNN dynamics.}
   \label{ScalingTC}
 \end{figure}

Maintaining the key spiking/temporal characteristics in the neural, synaptic, and learning processes is favorable because: 1)  the SNNs with time compression essentially attains pretty much the same dynamic behavior like before such that the classification performance would be also similar to the one under no time compression, i.e. no large performance degradation is expected when employing time compression; 2) the deployed learning rules need no modification and the same rules can effectively train the SNNs with time compression. 
Attaining the above goal entails proper scaling of  the time constants associated with these processes  as a function of the time compression ratio as shown in Fig.~\ref{ScalingTC}.



Without loss of generality, consider a decaying first order dynamics $\dot x(t)=-x(t)/\tau$  with time constant $\tau$. For digital hardware implementation, forward Euler discretization may be adopted to discretize the dynamics over time:

\begin{equation}
\label{Euler discretization}
\begin{array}{l}
X(t + \Delta t) = X(t)\left( {1 - \frac{{\Delta t}}{\tau }} \right)= X(t)\left( {1 - \frac{1}{{{\tau _{nom}}}}} \right)
\end{array}
\end{equation}
where $\Delta t$ is the discretization time stepsize and $\tau_{nom} = \tau/\Delta t$ is the normalized time constant used in digital hardware implementation. Now denote the target time compression ratio by $\gamma$ ($\gamma \geq 1)$. The discretization  stepsize with time compression is:  $\Delta t_c = \gamma\Delta t$, i.e. one time step of the time-compressed SNN  equals to  $\gamma$ time steps of the uncompressed SNN. 
Based on (\ref{Euler discretization}), discretizing the first order dynamics with time compression for one step gives:  
\begin{equation}
\label{DiscretizedFirstOrrderDynamics}
X(t + \Delta {t_c}) = X(t){\left( {1 - \frac{{1}}{\tau_{nom,c} }} \right)} = X(t){\left( {1 - \frac{{1}}{\tau_{nom} }} \right)^\gamma },
\end{equation}
where  $\tau _{nom,c}$ is the normalized time constant with compression. Linearly scaling $\tau _{nom,c}$  by ${\tau _{nom,c}}{\rm{ = }}\frac{{{\tau _{nom}}}}{\gamma }$ is equivalent to: $X(t + \Delta {t_c}){\rm{ \approx }}X(t)\left( {{\rm{1 - }}\frac{1}{{{\tau _{nom}}/\gamma }}} \right)$, which produces large errors when $\gamma \gg 1$. Instead,  we get an accurate $\tau_{nom,c}$ value according to: 
${\tau _{nom,c}} = \frac{1}{{{\rm{1 - }}{{\left( {1 - \frac{{1}}{\tau_{nom} }} \right)}^\gamma }}}$.

\subsection{Flexible Compression Ratios using Time Averaging}
Digital multipliers and dividers are costly in area and power dissipation. Normalized time constants in a digital SNN hardware accelerator  are typically set to a power of 2, i.e. $\tau_{nom} = 2^K$ such that the dynamics can be efficiently implemented by a shifter rather than expensive multipliers and dividers \cite{17}. However, it may be desirable to  choose a compression ratio and/or scale each time constant continuously in a wide integer range, e.g. within $\{1, 2, 3, ..., 16\}$. In this case, each scaled normalized time constant $\tau_{nom,c}$ may not be a power of 2. For example, when $\tau_{nom,c} = 10$,  $\tau_{nom,c}$ is far away from its two nearest powers of 2, namely 8 and 16.  Setting $\tau_{nom,c}$ to either of the two would lead to large errors. 

\begin{figure}[htpb]
  \centering
  \includegraphics[width=0.9\linewidth]{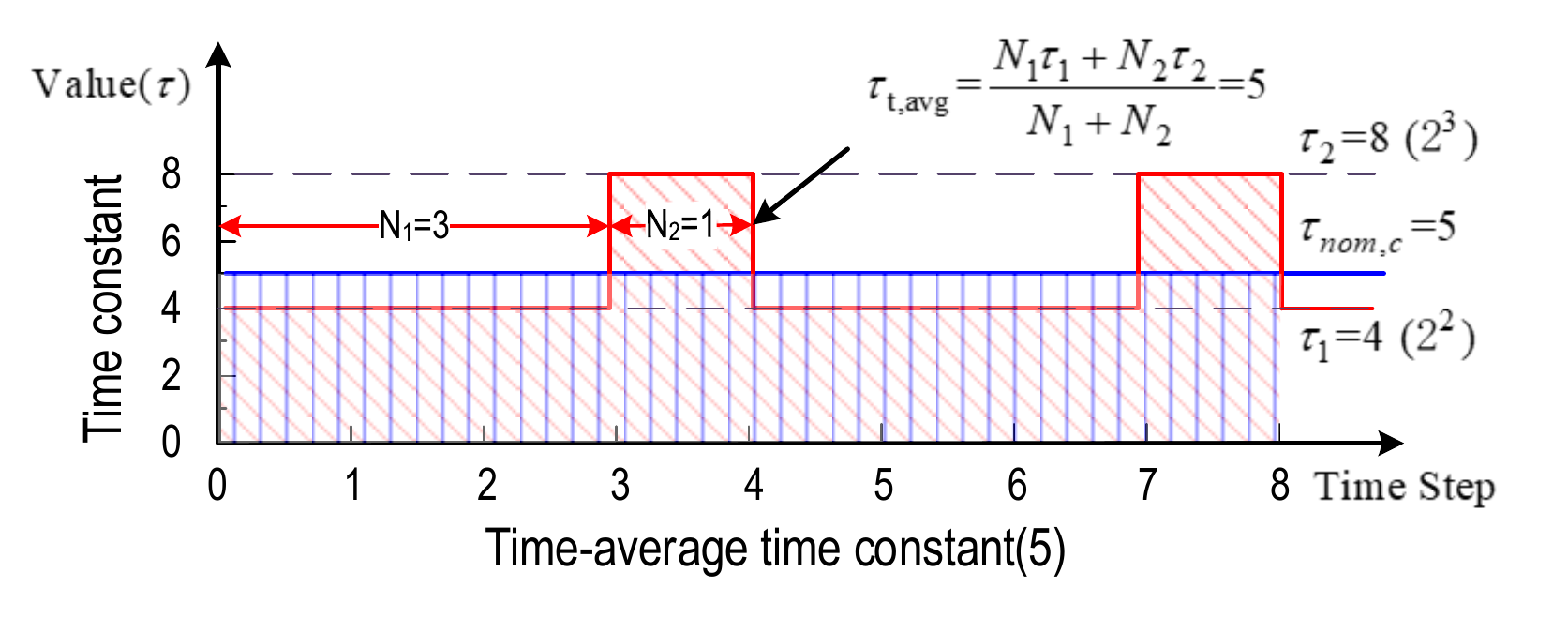}
  \caption{Time-averaged time constants: the realized averaged time constant is 5. }
  \label{Time_average}
\end{figure}

We propose a novel time averaging approach to address the above problem (Fig. \ref{Time_average}). For a given scaled normalized $\tau_{nom,c}$ , we find its two adjacent powers of 2: $2^{K_2}\leq \tau_{nom,c} \leq 2^{K_1}$ . We decay the targeted first order dynamics by toggling its scaled normalized time constant between two values: $2^{K_2}$ and $2^{K_1}$. Since each of them is a power of two, the corresponding decaying behavior can be efficiently realized using a shifter. The usage frequencies of  $2^{K_2}$ and $2^{K_1}$ are properly chosen such the time-averaged time constant  is equal to the desired $\tau_{nom,c}$. Fig. \ref{Time_average} shows how the time-averaged (normalized) time constant value of  5 is achieved by averaging between two compression ratios 4 and 8. 

\section{Proposed Input-and-Output Weighted (IOW) Spiking Neural Models}\label{sec:IOW}
Any given spiking neural model can be converted into its input-and-output (IOW) counterpart based on straightforward low-overhead modifications.  Without loss of generality, we consider conversion of two models:  the standard leaky integrate-and-fire (LIF) neuron model, which has been widely used in many SNNs including ones based on rating coding, and one of its variants for  supporting burst coding. 

\subsection{IOW Neurons based on Standard LIF Model}
The LIF model dynamics is \cite{12}:

\begin{equation}
\label{LIFDynamicRate}
  {\tau _m}\frac{{du}}{{dt}} =  - u(t) + RI(t),
\end{equation}
where $u(t)$ is the membrane potential, ${\tau _m}{\rm{ = }}RC$ is the membrane  time constant, and $I(t)$ is the  total received post-synaptic current given by: 

\begin{equation}
    \label{It}
    I(t) = \sum_i w_{i} \sum_f \alpha (t - t_i^{(f)}),
\end{equation}
where $w_{i}$ is the synaptic weight  from the pre-synaptic neuron $i$,  $\alpha(t) = \frac{q}{\tau_s} \exp\left(-\frac{t}{\tau_s} \right) H(t)$ for a  first order synaptic model with time constant $\tau_s$, $H(t)$ is the Heaviside step function,  and $q$ is the total charge injected into the post-synaptic neuron through a synapse of a weight of 1. In this work, we adopt a somewhat more complex second order model for improved performance. 

Once the membrane potential reaches the firing threshold $u_{th}$, an output spike is generated and the membrane potential is reset according to: 
\begin{equation}
    \label{Reset}
\mathop {\lim }\limits_{\delta  -  > 0;\delta  > 0} u({t^{(f)}} + \delta ) = u({t^{(f)}}) - {u_{th}},
\end{equation}
where $t^{(f)}$ is the firing time.

IOW LIF neurons shall process weighted input spikes because of time compression with the modified synaptic input:

\begin{equation}
    \label{ItIOWLIF}
    I(t) = \sum_i w_{i} \sum_f \omega_{spike, i}^f \alpha (t - t_i^{(f)}),
\end{equation}
where a weight $\omega_{spike, i}^f$ is introduced for each input spike. 

IOW LIF neurons shall also generate weighted output spikes. According to Fig.~\ref{WeightOutput}, we introduce a set of firing thresholds $\{u_{th}$, 2$u_{th}$, ... ,n$u_{th} \}$ with each being a multiple of the original threshold  $u_{th}$. At each time step $t$, an output spike is generated whenever the membrane potential reaches above any firing threshold from the set and the weight of the output spike is determined by the actual threshold crossed. For example, when $ k u_{th} \leq u(t) < (k+1) u_{th} $, the output spike weight is set to $k$. Upon firing, the membrane potential is reset according to:

\begin{equation}
\label{ResetIOWLIF}
\mathop {\lim }\limits_{\scriptstyle\delta  -  > 0\hfill\atop
\scriptstyle\delta  > 0\hfill} u({t^{(f)}} + \delta ) = \left\{ {\begin{array}{*{20}{l}}
{u({t^{(f)}}) - {u_{th}},}&{{u_{th}} \le u({t^{(f)}}) < {u_{th}}}\\
{u({t^{(f)}}) - 2{u_{th}},}&{2{u_{th}} \le u({t^{(f)}}) < 3{u_{th}}}\\
{...}&{...}\\
{u({t^{(f)}}) - n{u_{th}},}&{u({t^{(f)}}) \ge n{u_{th}}}
\end{array}} \right.
\end{equation}

\subsection{IOW Neurons based on Bursting LIF Model}
The LIF model for burst coding is also based on (\ref{LIFDynamicRate}) \cite{park2019fast}. A bursting function $g_i(t)$  is introduced to implement the bursting behavior per each presynaptic neuron $i$ \cite{park2019fast}:
\begin{equation}
\label{burstfunction}
g_i(t) = \left\{ {\begin{array}{*{20}{l}}
{\beta g_i(t - \Delta t)},\\
1,
\end{array}} \right.\begin{array}{*{20}{l}}
{\textrm{if}\ {\rm{ }}{E_i}(t - \Delta t) = 1} \\
{\textrm{otherwise}}
\end{array}
\end{equation}
where $\beta$ is a burst constant,  $E_{i}(t - \Delta t) =1 $ if the presynaptic neuron $i$ fired at the previous time step and otherwise $E_{i}(t - \Delta t) =0 $. We assume a zero-th order synaptic response model. 
Per input spikes from the presynaptic neuron $i$, the  firing  threshold voltage is modified from  $u_{th}$  to $g_i(t)u_{th}$ and the corresponding reset characteristic of the membrane potential after firing is: 
\begin{equation}
\label{ResetIOWBrust}
\mathop {\lim }\limits_{\delta  -  > 0;\delta  > 0} u({t^{(f)}} + \delta ) = u({t^{(f)}}) - g_i(t^{(f)}){u_{th}}. 
\end{equation}
Furthermore, the total post-synaptic current is: 
\begin{equation}
    \label{ItIOWBrust}
    I(t) = \sum_i w_{i} \sum_f g_i(t)\alpha (t - t_i^{(f)}). 
\end{equation}

To implement the IOW version of the LIF model with burst coding, we modify the burst function to:
\begin{equation}
\label{burstfunctionIOW}
g_i(t) = \left\{ {\begin{array}{*{20}{l}}
{\beta^{{\omega_{spike, i}(t)}} g(t - \Delta t)},\\
1,
\end{array}} \right.\begin{array}{*{20}{l}}
{\textrm{if } {E_i}(t - \Delta t) = 1}\\
{\textrm{otherwise}}
\end{array}
\end{equation}

Similar to the case of the IOW LIF model, we use a set of firing thresholds to determine the weight of each output spike and a behavior similar to   (\ref{ResetIOWLIF}) for reset. The only difference here is that the adopted set of firing thresholds are  $g_i(t)u_{th}$, $2g_i(t)u_{th}$, $\cdots$ ,$n g_i(t)u_{th}$ . 

\section{Time-Compressed SNN Accelerator Architectures}

\begin{figure}[htbp]
  \centering
  \includegraphics[width=0.9\linewidth]{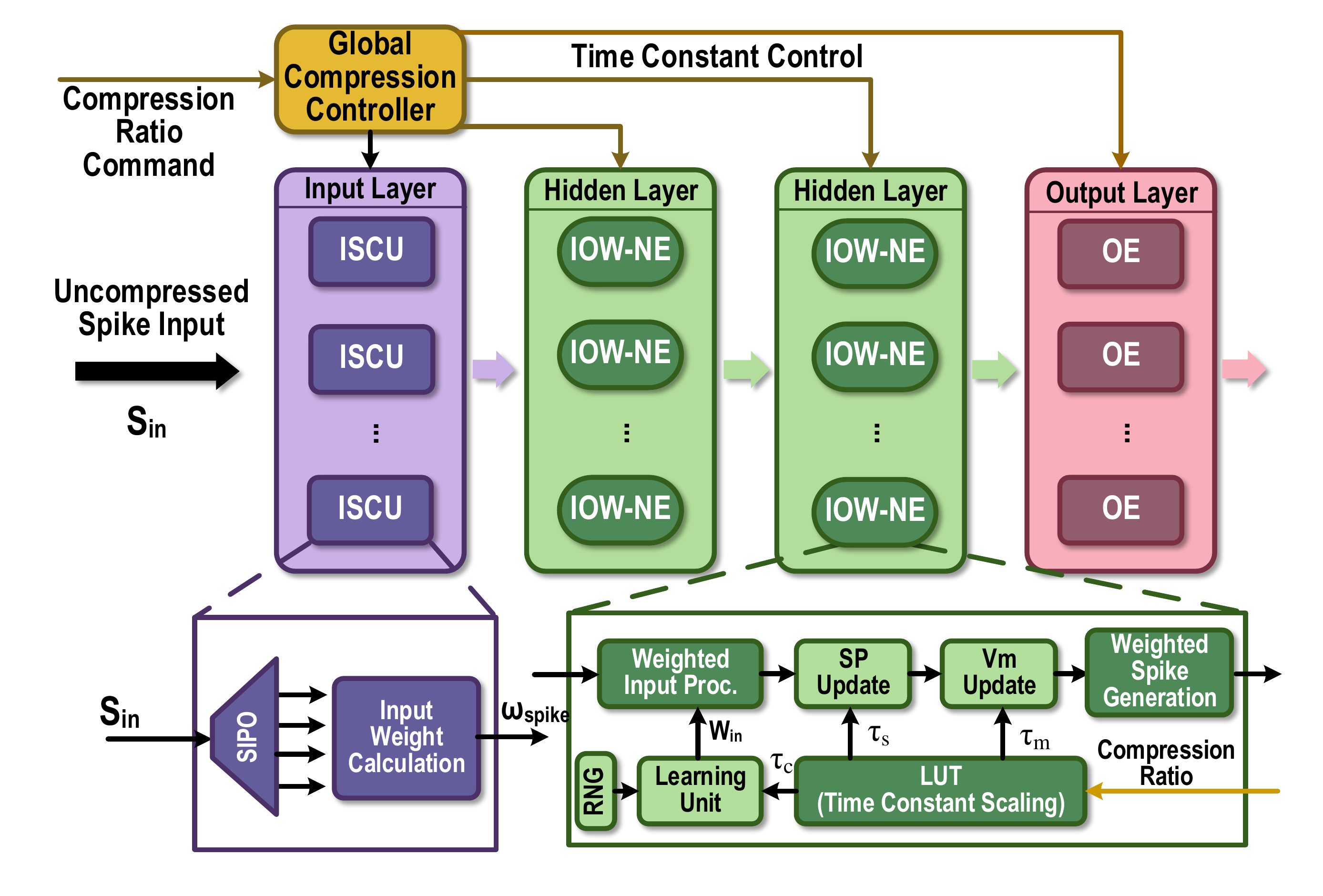}
  \caption{ Proposed time-compressed SNN architecture with programmable compression ratio (PTC-SNN).  ISCU: input spike compression unit; SIPO: serial-in and parallel-out; IOW-NE: input-output-weighted spiking neuron element; SP: synapse response; NE: regular binary-input-output neuron element; Vm: membrane potential. The LUT enables programmable scaling of time constants of the neuron/synaptic models and the learning unit.}
  \label{Architecture}
\end{figure}

The proposed time compression technique can be employed to support a fixed time compression ratio or user-programmable time compression ratio, leading to the time-compressed SNN (TC-SNN) and programmable time-compressed SNN (PTC-SNN) architectures, respectively. We describe the more general PTC-SNN architecture shown in Fig.~\ref{Architecture}. It can be adopted for any pre-designed SNN hardware accelerator for added programmable time compression. PTC-SNN introduces three streamlined additions and minor modifications to the embedded SNN accelerator to enable application and coding independent time compression. 

Based on the discussions presented in Section~2, firstly, a set of input-spike compression units (ISCUs), one for each input spike channel, are incorporated into the input layer of the SNN. ISCUs convert the raw binary input spike trains into the more compact weighted form with shortened time duration. A user-specified command sets the time compression ratio of all ISCUs through the Global Compression Controller. 
ISCUs compress the given spike channels without assuming sparsity of the input spike trains and can support large compression ratios. 
Secondly,  we introduce modest added hardware overhead to replace all original silicon spiking neurons  by their input-output-weighted neuron elements (IOW-NEs). 
Finally, all time constants in the SNN are scaled based on the time compression ratio. While an SNN may employ a large number of time constants, they can be all scaled in the same way, allowing use of one common simple programmable logic unit, i.e. the
Global Compression Controller for scaling all time constants according to  a user-specified compression ratio command. 

\begin{figure}[htbp]
  \centering
  \includegraphics[width=\linewidth]{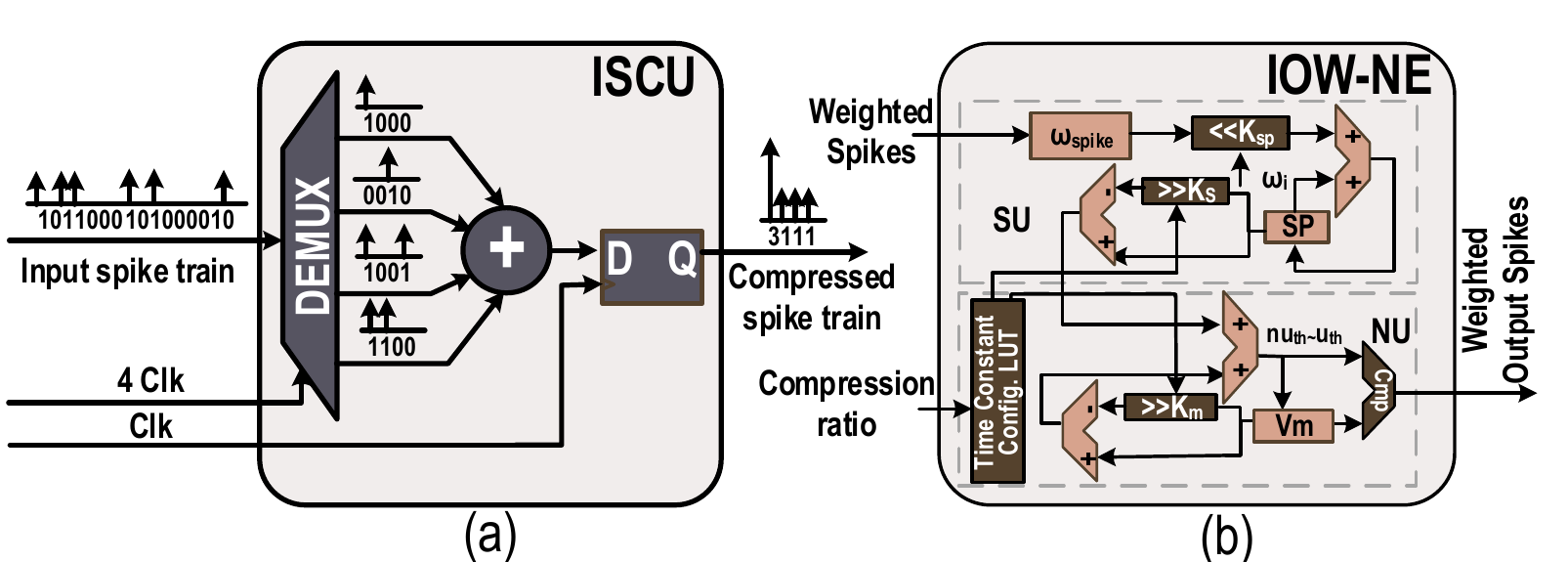}
  \caption{(a) ISCU with 4:1 time compression, and (b) LIF IOW neuron: SU - synaptic unit, NU - neural unit.}
  \label{FIG_ISCU_IOW}
\end{figure}

 \textbf{[Input Spike Compression Unit (ISCU)]}
Each input spike channel is compressed by one low-cost ISCU according to the user-specified compression ratio $N_{cmp}$. When each uncompressed spike input channel is fed by a single binary serial input, a demultiplexer is utilized in the ISCU to perform the reconfigurable serial-in and parallel-out (SIPO) operation to convert the serial input into $N_{cmp}$ parallel outputs, as shown in Fig.~\ref{FIG_ISCU_IOW}(a). If the input spike channel is supplied by parallel spike data, the SIPO operation is skipped. During each clock cycle, the $N_{cmp}$ bits of the parallel outputs are added by an adder, which effectively combines these spikes into a single weighted spike with a weight value set by the output of the adder.
No spike count loss is resulted as the sum of spike weights is same as the total number of binary spikes in the raw spike input train. The global temporal spike distribution of the input spike train is preserved up to the temporal resolution of the compressed spike train. 

\textbf{[Input-Output-Weighted (IOW) Neuron Elements]} 
We discuss efficient hardware realization of the IOW spiking neural models (Section~3). The IOW neuron element (IOW-NE)  is shown in  Fig. \ref{FIG_ISCU_IOW}(b), which consist of a synaptic unit (SU), a neural unit (NU), and a time constant configuration module, described later. SU realizes a discretized version of (\ref{ItIOWLIF}). As in many practical implementations of hardware SNNs, each $\omega_i$  is constrained to be in the form of $2^K$. The product of $\omega_{spike,i} \cdot \omega_i$ is efficiently realized by left shifting $\omega_{spike,ki}$ by K bits. NU performs membrane potential $u(t)$ update based on discretization of (\ref{LIFDynamicRate}) and reset behavior (\ref{ResetIOWLIF}).  NU generates a weighted output spike when $u(t)$ is above certain threshold in the firing threshold set $u_{th}$, $2u_{th}$, $\cdots$.
   

 The design of IOW LIF neurons with burst coding is almost identical to that of the IOW LIF neurons except for the following differences. We add a LUT to store the set of firing thresholds  $\{ g_i(t) u_{th}$, $2g_i(t) u_{th}$, $\cdots \}$,  which are calculated based on  (\ref{burstfunctionIOW}). Because  $g_i(t)u_{th}$ might not be in the form of $2^K$, a multiplier is used to compute the product $g(t) \cdot u_{th} \cdot \omega_i \cdot \omega_{spike, i}$.

\begin{table*}[h]
\centering
\setlength{\tabcolsep}{2.5pt}
\caption{Comparison of the baseline and TC-SNN accelerators with IW/IOW LIF neurons based on TI46 Speech Corpus. }
\label{Performance_TI46_Speech_Corpus}
\begin{tabular}{lllllllllll} 
\hline
\begin{tabular}[c]{@{}l@{}} Compres-\\sion ratio \end{tabular} & \begin{tabular}[c]{@{}l@{}} Neuron\\model \end{tabular} & \begin{tabular}[c]{@{}l@{}} Accuracy\\\end{tabular} & LUT   & FF    & \begin{tabular}[c]{@{}l@{}} Power(W)\\@50MHz \end{tabular} & \begin{tabular}[c]{@{}l@{}} Runtime(s)\\(Normalized\\Runtime) \end{tabular} & \begin{tabular}[c]{@{}l@{}} Runtime\\Speedup \end{tabular} & \begin{tabular}[c]{@{}l@{}} Energy(J)\\(Normalized\\Energy) \end{tabular} & \begin{tabular}[c]{@{}l@{}} Energy\\reduction\\ratio\end{tabular} & \begin{tabular}[c]{@{}l@{}} Normalized\\ATEL \end{tabular}  \\ 
\hline
baseline                                                       & LIF                                                     & 96.15\%                                             & 57326 & 18200 & 0.073                                                      & 1.991(100\%)                                                                & 1.00x                                                      & 0.145(100\%)                                                              & 1.00x                                                             & 100\%                                                       \\
2:1                                                            & IW-LIF                                                  & 96.15\%                                             & 58497 & 18460 & 0.077                                                      & 0.995(49.97\%)                                                              & 2.00x                                                      & 0.077(52.71\%)                                                            & 1.88x                                                             & 26.68\%                                                     \\
2:1                                                            & IOW-LIF                                                 & 96.15\%                                             & 60096 & 18532 & 0.086                                                      & 0.995(49.97\%)                                                              & 2.00x                                                      & 0.086(58.87\%)                                                            & 1.69x                                                             & 30.72\%                                                     \\
3:1                                                            & IW-LIF                                                  & 92.31\%                                             & 58762 & 18782 & 0.080                                                      & 0.664(33.35\%)                                                              & 3.00x                                                      & 0.053(36.55\%)                                                            & 2.74x                                                             & 24.98\%                                                     \\
3:1                                                            & IOW-LIF                                                 & 92.31\%                                             & 61162 & 18799 & 0.092                                                      & 0.664(33.35\%)                                                              & 3.00x                                                      & 0.061(42.03\%)                                                            & 2.38x                                                             & 29.74\%                                                     \\
4:1                                                            & IW-LIF                                                  & 92.31\%                                             & 58910 & 18753 & 0.081                                                      & 0.499(25.06\%)                                                              & 3.99x                                                      & 0.036(27.81\%)                                                            & 4.03x                                                             & 14.31\%                                                     \\
4:1                                                            & IOW-LIF                                                 & 92.31\%                                             & 61313 & 18923 & 0.095                                                      & 0.499(25.06\%)                                                              & 3.99x                                                      & 0.047(32.62\%)                                                            & 3.09x                                                             & 17.40\%                                                     \\
8:1                                                            & IW-LIF                                                  & 80.77\%                                             & 59210 & 19087 & 0.083                                                      & 0.248(12.46\%)                                                              & 8.03x                                                      & 0.021(14.16\%)                                                            & 6.90x                                                             & 9.12\%                                                      \\
8:1                                                            & IOW-LIF                                                 & 86.54\%                                             & 62548 & 19098 & 0.099                                                      & 0.248(12.46\%)                                                              & 8.03x                                                      & 0.025(16.89\%)                                                            & 5.80x                                                             & 7.98\%                                                      \\
16:1                                                           & IW-LIF                                                  & 69.23\%                                             & 59400 & 20000 & 0.117                                                      & 0.125(6.28\%)                                                               & 15.93x                                                     & 0.015(10.06\%)                                                            & 9.67x                                                             & 5.28\%                                                      \\
16:1                                                           & IOW-LIF                                                 & 80.77\%                                             & 65349 & 20808 & 0.134                                                      & 0.125(6.28\%)                                                               & 15.93x                                                     & 0.017(11.52\%)                                                            & 8.53x                                                             & 4.12\%                                                      \\
\hline
\end{tabular}
\end{table*}

\section{Experimental Evaluations}
The proposed time-compressed SNN (TC-SNN) architecture with a  fixed compression ratio and  the more general programmable PTC-SNN architecture with user-programmable compression ratio can be adopted to re-design any given digital SNN accelerator to a time-compressed SNN accelerator with low additional design overhead in a highly streamlined manner. For demonstration purpose, we show how an existing liquid state machine (LSM) SNN accelerator can be re-designed to a TC-SNN and PTC-SNN on a Xilinx Zynq-7000 FPGA. The LSM is a recurrent spiking neural network model. With its spatio-temporal computing power, it has demonstrated promising performances for various applications \cite{15}. 

Three speech/image recognition datasets are adopted for benchmarking. The first dataset is a subset of the TI46 speech corpus \cite{10} and consists of 260 isolated spoken English letters recorded by a single speaker. The time domain speech examples are pre-processed by the Lyon’s passive ear model \cite{13} and transformed to 78 channel spike trains using the BSA spike encoding algorithm \cite{14}. The second one is the CityScape  dataset \cite{11} which contains 18 classes of 1,080 images of semantic urban scenes taken in several European cities. Each image is segmented and remapped into a size of 15 $\times$ 15, are then converted to 225 Poisson spike trains with the mean firing rate proportional to the corresponding pixel intensity. The third one is a subset of N-TIDIGITS18  speech dataset \cite{anumula2018feature} which is obtained by playing the audio files from the TIDIGITS dataset to a CochleaAMS1b sensor. This dataset contains 10 classes of single digits (the digits “0” to “9”). There are  111 male and 114 female speakers in the dataset and 2,250 training and 2,250 testing examples. For the first two datasets, we adopt 80\% examples for training and the remaining 20\% for testing. The three datasets present two different types tasks, i.e. speech vs. image classification, and are based on three different raw input encoding schemes, i.e. the BSA encoding, Poisson-based rate coding, and CochleaAMS1b sensor based coding. Therefore, they are well suited for testing the generality of the proposed time compression.

The baseline LSM FPGA accelerator (without compression) we built in this paper is based on the standard LIF model, and consists of an input layer, a recurrent reservoir, and a readout layer. The number of input neurons is set by the number of the input spike trains, which is 78, 225 and 64, respectively for the TI46 dataset, CityScape dataset, and N-TIDIGITS18 dataset, respectively. The reservoir has 135 neurons for  the TI46 and CityScape datasets and 300 neurons for the N-TIDIGITS18 dataset, respectively.
The reservoir neurons are fully connected to the readout neurons. All readout synapses are plastic and trained using the supervised spike-dependent training algorithm in \cite{17}.  
The power consumption of various FPGA accelerators is measured using the Xilinx Power Analyzer (XPA) tool and their recognition performances are measured from the FPGA board.

\begin{table*}
\setlength{\tabcolsep}{2.5pt}
\caption{Comparison of the baseline and TC-SNN accelerators with IOW LIF neurons based on the CityScape image dataset. }
\label{Performance_CityScape}
\centering
\begin{tabular}{lllllllllll} 
\hline
\begin{tabular}[c]{@{}l@{}} Compres-\\sion ratio \end{tabular} & \begin{tabular}[c]{@{}l@{}} Neuron\\model \end{tabular} & Accuracy & LUT   & FF    & \begin{tabular}[c]{@{}l@{}} Power(W)\\@50MHz \end{tabular} & \begin{tabular}[c]{@{}l@{}} Runtime(s)\\(Normalized\\Runtime) \end{tabular} & \begin{tabular}[c]{@{}l@{}} Runtime\\Speedup \end{tabular} & \begin{tabular}[c]{@{}l@{}} Energy(J)\\(Normalized\\Energy) \end{tabular} & \begin{tabular}[c]{@{}l@{}} Energy\\reduction\\ratio \end{tabular} & \begin{tabular}[c]{@{}l@{}} Normalized\\ATEL \end{tabular}  \\ 
\hline
baseline                                                       & LIF                                                     & 99.07\%  & 57017 & 16373 & 0.074                                                     & 1.497(100\%)                                                                & 1.00x                                                      & 0.111(100\%)                                                              & 1.00x                                                              & 100\%                                                       \\
2:1                                                            & IOW-LIF                                                 & 99.07\%  & 58826 & 17294 & 0.078                                                     & 0.749(50.03\%)                                                              & 2.00x                                                      & 0.058(52.25\%)                                                            & 1.91x                                                              & 27.31\%                                                     \\
3:1                                                            & IOW-LIF                                                 & 97.69\%  & 58895 & 17506 & 0.113                                                     & 0.499(33.33\%)                                                              & 3.00x                                                      & 0.056(50.45\%)                                                            & 1.98x                                                              & 43.72\%                                                     \\
4:1                                                            & IOW-LIF                                                 & 97.69\%  & 59276 & 17374 & 0.082                                                     & 0.375(25.05\%)                                                              & 3.99x                                                      & 0.031(27.93\%)                                                            & 3.58x                                                              & 18.00\%                                                     \\
8:1                                                            & IOW-LIF                                                 & 95.37\%  & 61254 & 19322 & 0.092                                                     & 0.189(12.63\%)                                                              & 7.92x                                                      & 0.017(15.32\%)                                                            & 6.53x                                                              & 10.73\%                                                     \\
16:1                                                           & IOW-LIF                                                 & 94.91\%  & 66350 & 21618 & 0.079                                                     & 0.096(6.41\%)                                                               & 15.59x                                                     & 0.008(7.21\%)                                                             & 13.88x                                                             & 2.84\%                                                      \\
\hline
\end{tabular}
\end{table*} 
\subsection{Reservoir responses of the LSMs}

\begin{table*}
\setlength{\tabcolsep}{2.5pt}
\caption{Comparison of the baseline and TC-SNN accelerators with IOW LIF neurons based on the NTIDIGITS18 dataset. }
\label{Performance_NTIDIGITS18}
\centering
\begin{tabular}{lllllllllll} 
\hline
\begin{tabular}[c]{@{}l@{}} Compres-\\sion ratio \end{tabular} & \begin{tabular}[c]{@{}l@{}} Neuron\\model \end{tabular} & Accuracy & LUT    & FF    & \begin{tabular}[c]{@{}l@{}} Power(W)\\@50MHz \end{tabular} & \begin{tabular}[c]{@{}l@{}} Runtime(s)\\(Normalized\\Runtime) \end{tabular} & \begin{tabular}[c]{@{}l@{}} Runtime\\Speedup \end{tabular} & \begin{tabular}[c]{@{}l@{}} Energy(J)\\(Normalized\\Energy) \end{tabular} & \begin{tabular}[c]{@{}l@{}} Energy\\reduction\\ratio \end{tabular} & \begin{tabular}[c]{@{}l@{}} Normalized\\ATEL \end{tabular}  \\ 
\hline
Baseline                                                       & LIF                                                     & 83.63\%  & 106263 & 25778 & 0.116                                                    & 424.61(100\%)                                                                      & 1.00x                                                      & 49.255(100\%)                                                             & 1.00x                                                              & 100\%                                                       \\
2:1                                                            & IOW-LIF                                                 & 82.82\%  & 111688 & 26070 & 0.110                                                    & 212.31(50.00\%)                                                                      & 2.00x                                                      & 23.354(47.41\%)                                                           & 2.11x                                                              & 26.04\%                                                     \\
3:1                                                            & IOW-LIF                                                 & 82.22\%  & 124756 & 28364 & 0.112                                                    & 141.50(33.32\%)                                                                      & 3.00x                                                      & 15.848(32.18\%)                                                           & 3.11x                                                              & 13.58\%                                                     \\
4:1                                                            & IOW-LIF                                                 & 81.91\%  & 112224 & 26158 & 0.113                                                    & 106.87(25.17\%)                                                                      & 3.97x                                                      & 12.076(24.52\%)                                                           & 4.08x                                                              & 7.17\%                                                      \\
8:1                                                            & IOW-LIF                                                 & 80.91\%  & 131614 & 28934 & 0.158                                                    & 53.61(12.63\%)                                                                       & 7.92x                                                      & 8.470(17.20\%)                                                            & 5.82x                                                              & 3.10\%                                                      \\
16:1                                                           & IOW-LIF                                                 & 74.54\%  & 128094 & 34707 & 0.174                                                    & 27.17(6.40\%)                                                                       & 15.63x                                                     & 4.728(9.60\%)                                                             & 10.42x                                                             & 1.16\%                                                      \\
\hline
\end{tabular}
\end{table*}

\begin{figure}[htbp]
  \centering
	\includegraphics[width=0.15\textwidth]{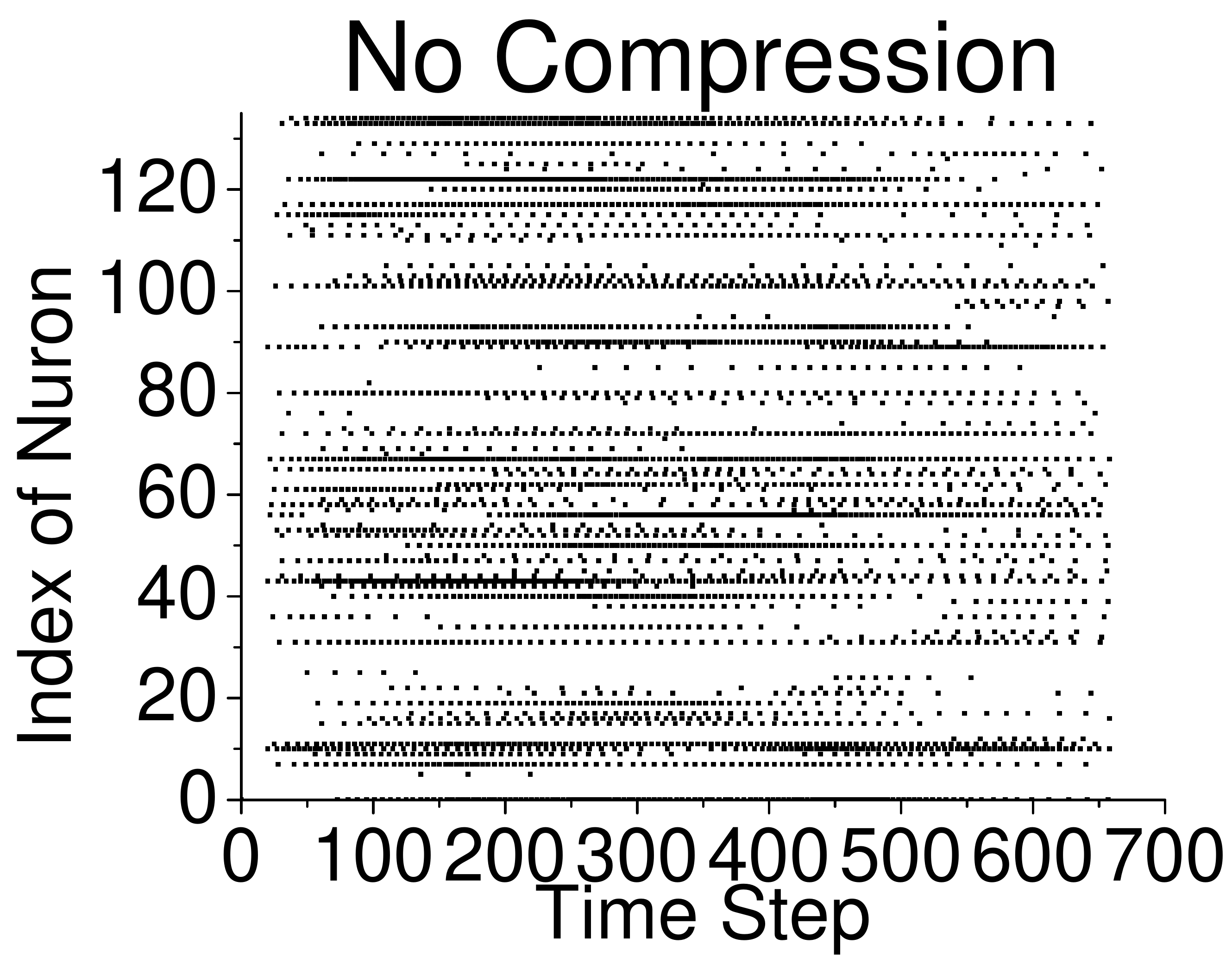}
	\includegraphics[width=0.15\textwidth]{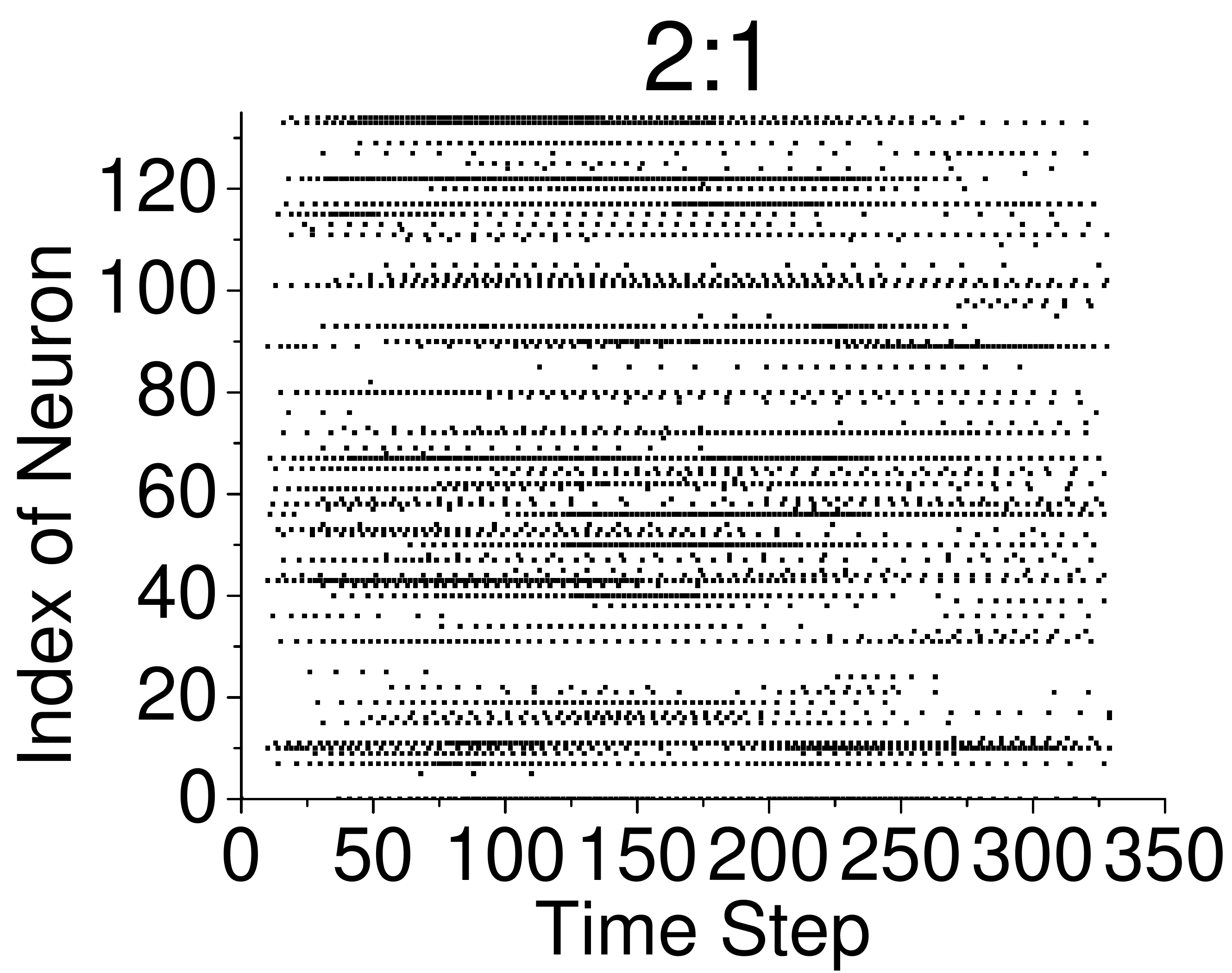}
	\includegraphics[width=0.15\textwidth]{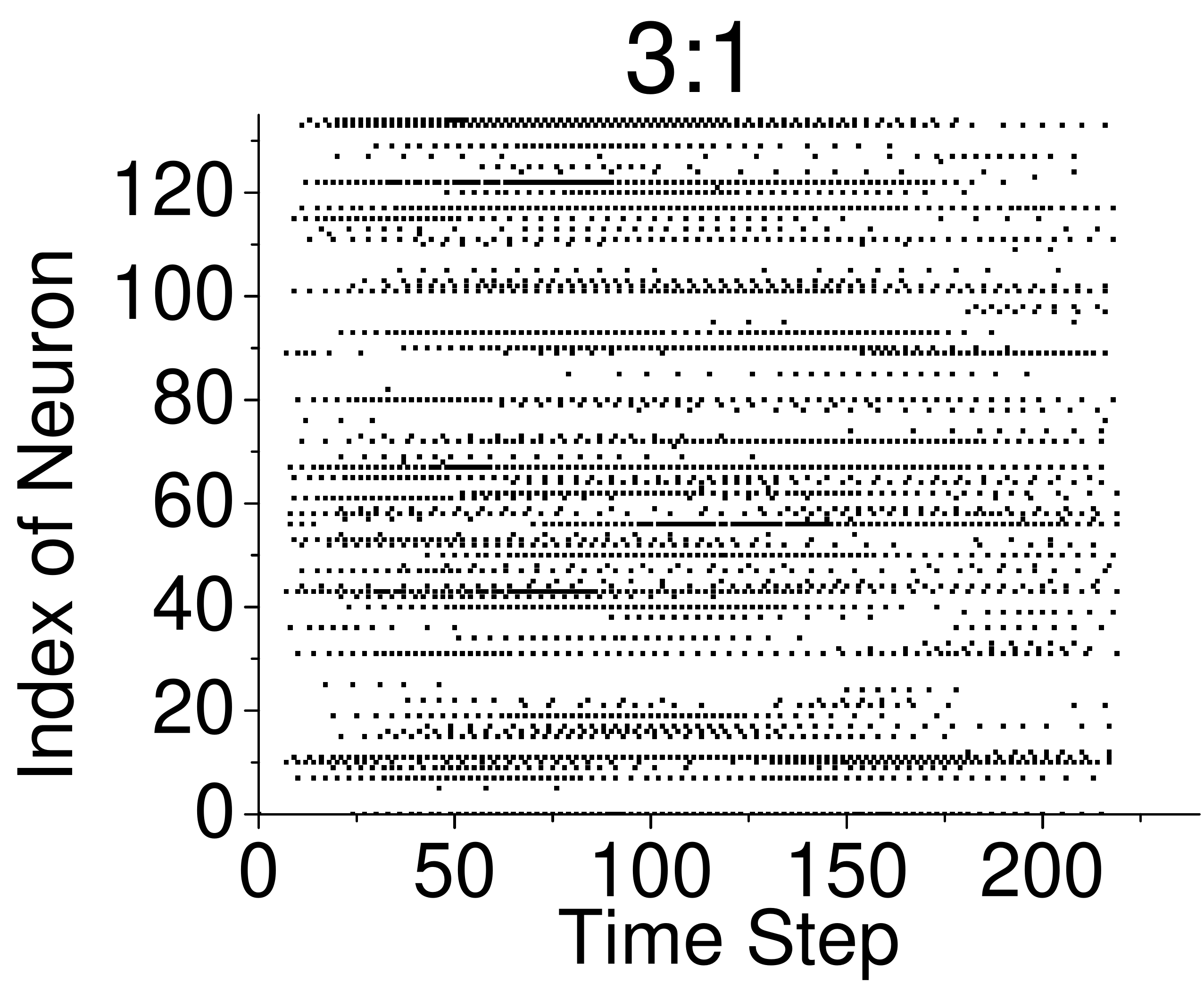}
	\includegraphics[width=0.15\textwidth]{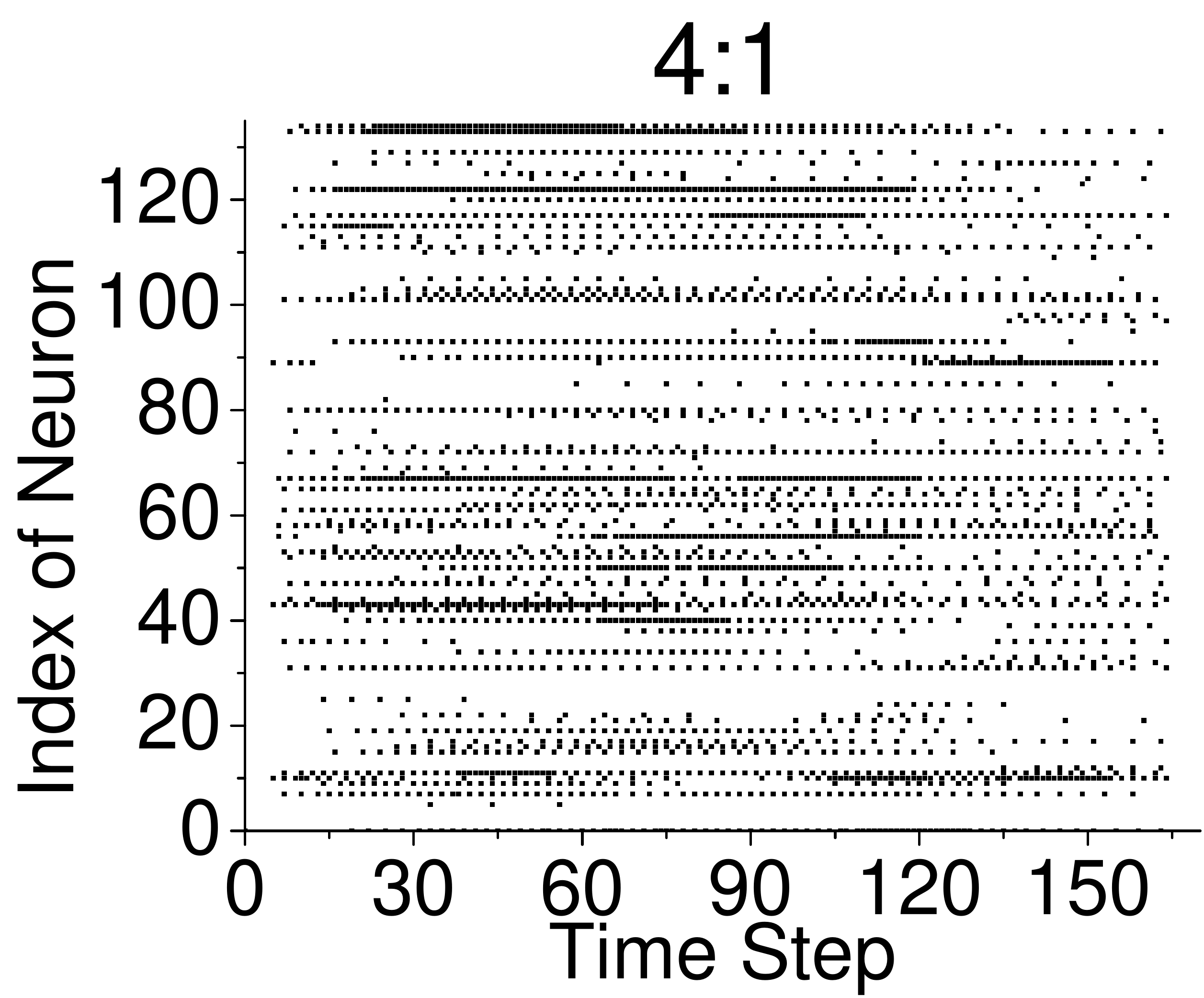}
	\includegraphics[width=0.15\textwidth]{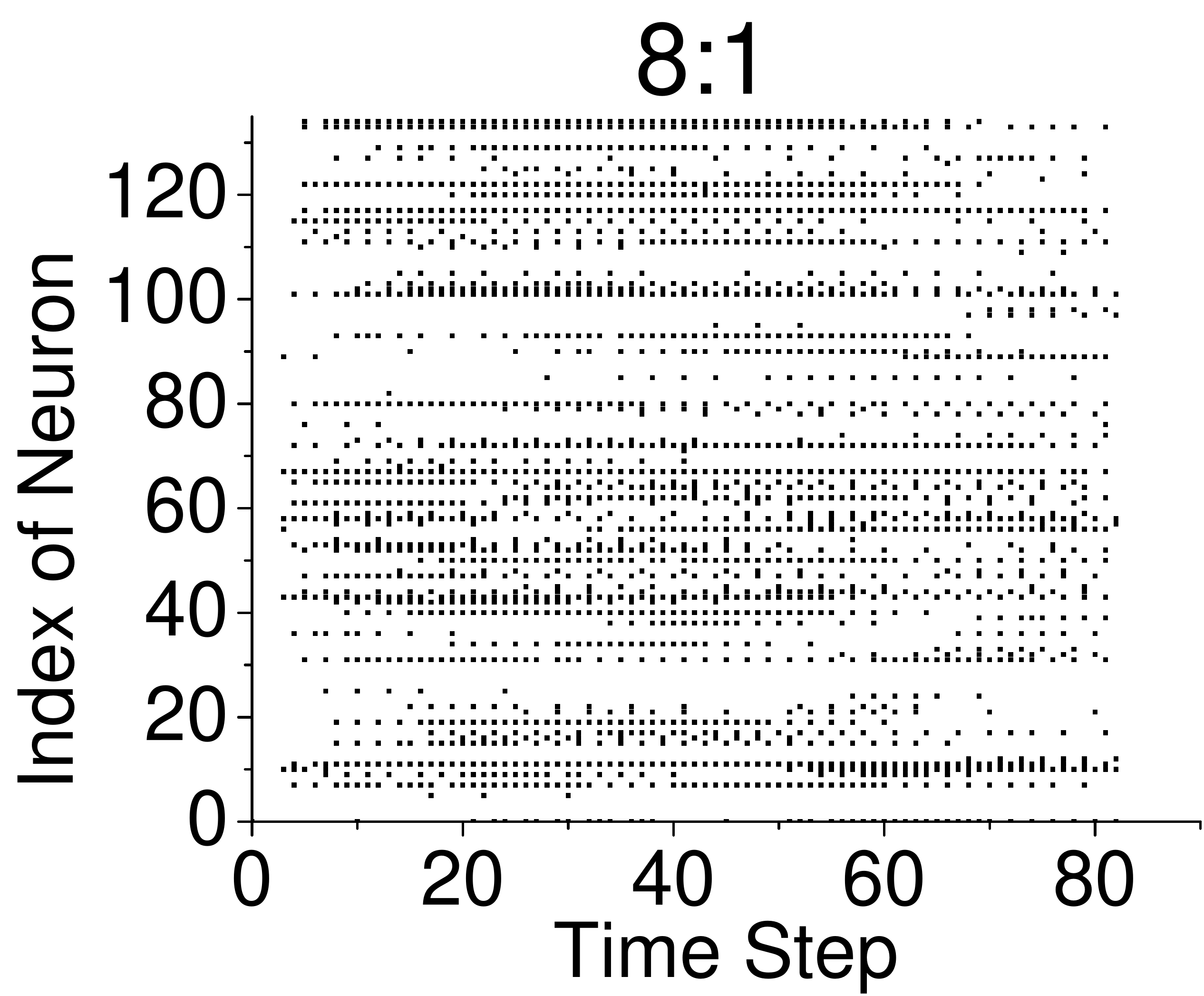}
	\includegraphics[width=0.15\textwidth]{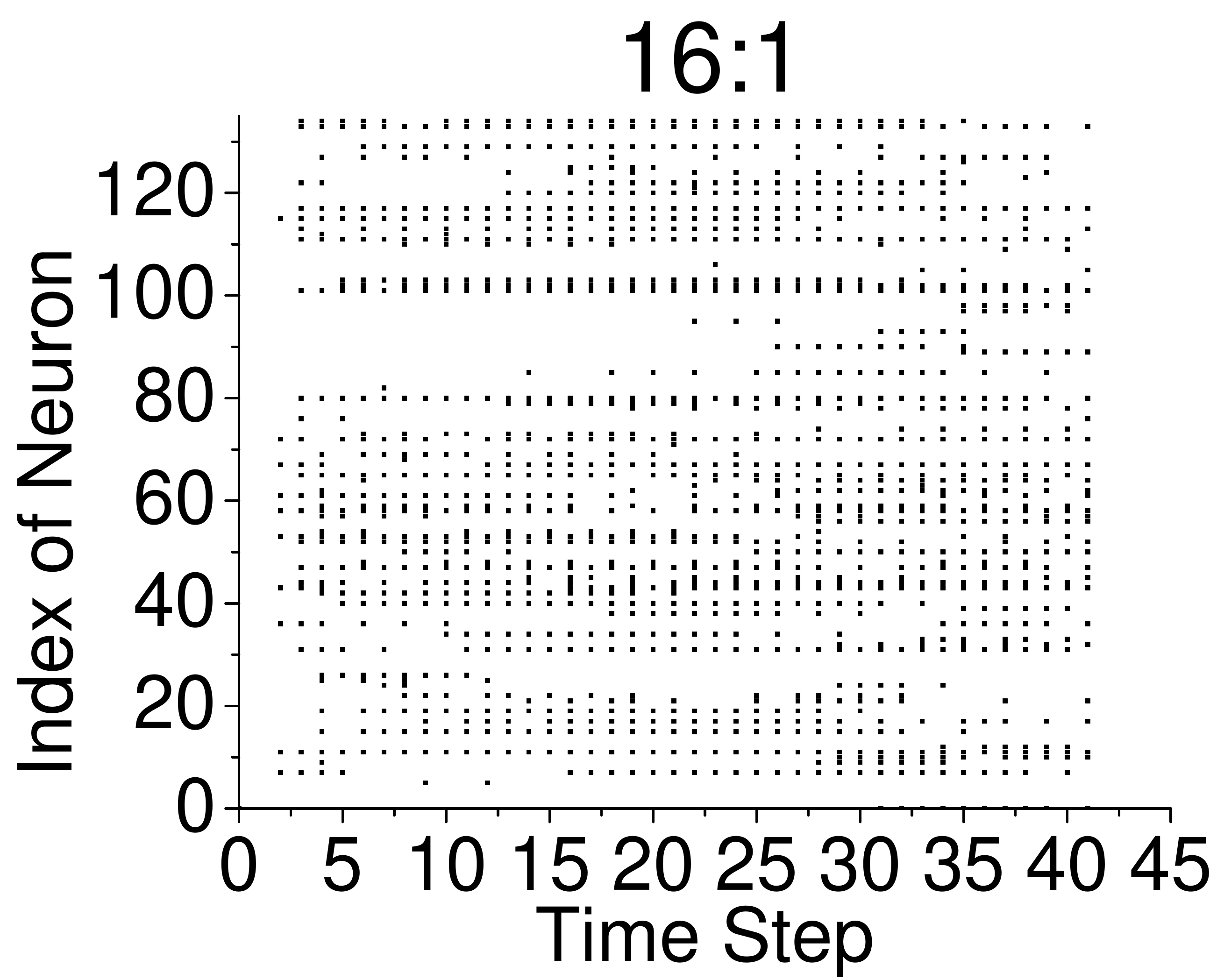}
	\caption{Reservoir response vs. compression ratio. }
	\label{Reservoir_response}
\end{figure}

We plot the raster plots of the reservoir IOW-LIF neurons when the input speech example is the letter “A” from the TI46 Speech Corpus to examine the impact of time compression in Fig. \ref{Reservoir_response}.
It is fascinating to observe that when the compression ratio is between 2:1 to 4:1, the reservoir response in terms of both total spike count and spatio-temporal spike distribution changes little from the one without compression. When the compression ratio increases to the very large values of 8:1 and 16:1, 
the original spatio-temporal spike distribution is still largely preserved. 
This is consistent to the decent recognition performance achieved at 8:1 and 16:1 compression ratios presented next. 

\subsection{Performances of TC-SNNs with IOW LIF Neurons}
For the three datasets mentioned, we design a baseline LSM SNN without time compression and five time-compressed SNNs (TC-SNNs) with IOW LIF neurons and a fixed time compression ratio from 2:1 to 16:1, all clocked  at 50MHz. 

For the TI46 speech dataset~\cite{10}, the runtime and energy dissipation of each accelerator expended on 350 training epochs of a batch of 208 randomly selected examples are measured.  We compare the inference accuracy, hardware overhead measured by FPGA lookup (LUT) and flip-flop (FF) utilization, power, runtime, and energy of all six accelerators in Table \ref{Performance_TI46_Speech_Corpus}. To show the benefit of producing weighted output spikes, we create a new input-weighted (IW) LIF model which differs from the IOW LIF model in that the IW model generates binary output spikes. We redesign the five TC-SNN accelerators using IW LIF neurons and compare them with their IOW counterparts in  Table \ref{Performance_TI46_Speech_Corpus}. With large compression ratios the IOW accelerators significantly outperform their IW counterparts on classification accuracy. For example, the IOW accelerator improves accuracy from  69.23\% to 80.77\% with a compression ratio of 16:1. 


The power/hardware overhead of the TC-SNN accelerators with IOW LIF neurons only increases modestly with the time compression ratio. 
Over a very wide range of compression ratio, the runtime is linearly scaled with the compression ratio while the energy is scaled almost linearly.  For example, 2:1 compression speeds up the runtime by 2$\times$, reduces the energy by 1.69$\times$, retaining the same classification accuracy of 96.15\% without degradation. With 4:1 compression, the runtime is sped up by 3.99$\times$, the energy is reduced by 3.09$\times$, and the classification accuracy is as high as 92.31\%. With a large 16:1 compression ratio, the runtime and energy are reduced significantly by 15.93$\times$ and 8.53$\times$, respectively, and the accuracy is 80.77\%.

To jointly evaluate the tradeoffs between hardware area, runtime, energy, and loss of accuracy, we define a figure of merit (FOM) ATEL as: ATEL = Area $\times$ Time $\times$ Energy $\times$ Loss, where each metric is normalized with respect to the baseline (no compression), and Loss = (100\% - Classification Accuracy). Here the hardware area is evaluated by Flop count + 2*LUT count as suggested by Xilinx. Table \ref{Performance_TI46_Speech_Corpus} shows that as the compression ratio increases from 1:1 to 16:1, the ATEL of the TC-SNNs with IOW LIF neurons favorably drops from 100\% to 4.12\%, a nearly 25-fold reduction.

We evaluate the proposed architectures using the CityScape image recognition dataset\cite{11} and N-TIDIGITS18 dataset\cite{anumula2018feature} in a similar way. The results for the CityScape datset are reported in Table \ref{Performance_CityScape}, for which the runtime and energy dissipation of each accelerator are measured for 350 training epochs of a batch of 864 randomly selected examples. Since the proposed compression is application independent, the TC-SNN architectures can be applied to this image recognition task without any modification. Large runtime and energy reductions similar to the ones for the TI46 dataset are achieved by the proposed time compression while the degradation of classification accuracy is more graceful. The TC-SNN with 8:1 compression reduces the runtime and energy dissipation by 7.92$\times$ and 6.53$\times$, respectively while the accuracy only drops to 95.37\%. The figure of merit ATEL improves from 100\% to 2.84\% (35$\times$ improvement) when the TC-SNN runs with 16:1 compression. The results on the N-TIDIGITS18 dataset are in Table \ref{Performance_NTIDIGITS18}, for which the runtime and energy dissipation of each accelerator are measured for 350 training epochs of a batch of 2,250 training samples. Again, large runtime and energy reductions are achieved by the proposed time compression. The TC-SNN with 8:1 compression ratio reduces the runtime and energy dissipation by 7.92$\times$ and 5.82$\times$ respectively while the accuracy only drop from 83.63\% to 80.91\%.

Clearly, the proposed compression architectures can linearly scale the runtime, and hence dramatically reduce the decision latency, and energy dissipation without significant accuracy degradation at low compression ratios, e.g. up to 4:1. Applying an aggressively large compression ratio can produce huge energy and runtime reduction while the degraded performance may be still acceptable for practical applications. The supported large range of compression ratio offers the user great flexibility in targeting an appropriate performance/overhead tradeoff for a given application.

\subsection{Performances of TC-SNNs with  Bursting Coding}
 We redesign our TC-SNN accelerators using bursting IOW LIF models to support burst coding \cite{park2019fast}  and compare their performances with the baseline on the TI46 speech dataset in Table \ref{Performance_TI46_bursting}.  Once again, the proposed time compression leads to large  runtime and energy reductions and the degradation of classification accuracy is graceful. 
 The additional hardware cost for supporting bursting coding is somewhat increased but still rather moderate. 

\begin{table*}
\setlength{\tabcolsep}{2.5pt}
\caption{Comparison of the baseline and TC-SNN accelerators with burst coding on the TI46 Speech Corpus. }
\label{Performance_TI46_bursting}
\centering
\begin{tabular}{lllllllllll} 
\hline
\begin{tabular}[c]{@{}l@{}} Compres-\\sion ratio \end{tabular} & \begin{tabular}[c]{@{}l@{}} Neuron\\model \end{tabular} & Accuracy & LUT    & FF    & \begin{tabular}[c]{@{}l@{}} Power(W)\\@50MHz \end{tabular} & \begin{tabular}[c]{@{}l@{}} Runtime(s)\\(Normalized\\Runtime) \end{tabular} & \begin{tabular}[c]{@{}l@{}} Runtime\\Speedup \end{tabular} & \begin{tabular}[c]{@{}l@{}} Energy(J)\\(Normalized\\Energy) \end{tabular} & \begin{tabular}[c]{@{}l@{}} Energy\\reduction\\ratio \end{tabular} & \begin{tabular}[c]{@{}l@{}} Normalized\\ATEL \end{tabular}  \\ 
\hline
baseline                                                       & LIF                                                     & 98.08\%  & 92052  & 62390 & 0.240                                                     & 2.527(100\%)                                                                & 1.00x                                                      & 0.606(100\%)                                                              & 1.00x                                                              & 100\%                                                       \\
2:1                                                            & IOW-LIF                                                 & 92.31\%  & 107263 & 64845 & 0.163                                                     & 1.266(50.10\%)                                                              & 2.00x                                                      & 0.206(33.99\%)                                                            & 2.94x                                                              & 77.38\%                                                     \\
3:1                                                            & IOW-LIF                                                 & 92.31\%  & 124881 & 67343 & 0.168                                                     & 0.946(37.44\%)                                                              & 2.67x                                                      & 0.158(26.07\%)                                                            & 3.82x                                                              & 50.55\%                                                     \\
4:1                                                            & IOW-LIF                                                 & 92.31\%  & 102362 & 61332 & 0.172                                                     & 0.637(25.21\%)                                                              & 3.97x                                                      & 0.110(18.15\%)                                                            & 5.54x                                                              & 19.68\%                                                     \\
8:1                                                            & IOW-LIF                                                 & 88.46\%  & 121183 & 64481 & 0.212                                                     & 0.318(12.58\%)                                                              & 7.95x                                                      & 0.067(11.06\%)                                                            & 9.00x                                                              & 10.47\%                                                     \\
16:1                                                           & IOW-LIF                                                 & 80.77\%  & 132055 & 72508 & 0.289                                                     & 0.163(6.45\%)                                                               & 15.50x                                                     & 0.047(7.76\%)                                                             & 12.87x                                                             & 6.85\%                                                      \\
\hline
\end{tabular}
\end{table*}

\subsection{Performances of PTC-SNNs with Reconfigurable Compression Ratio}
We also design a time-compressed SNN (PTC-SNN) accelerator supporting programmable ratio ranging from 2:1 to 16:1 and evaluate it using the TI46 dataset in Table~\ref{Performance_TI46_PTC}. The LUT and FF utilizations of PTC-SNN are 7,4742 and 2,1391, respectively. The overall hardware area overhead stays constant with the programmable compression ratio, which is only 12.78\% more than that of the TC-SNN accelerator with a fixed 16:1 compression ratio. 
The runtime and accuracy of the PTC-SNN are identical to those of the corresponding TC-SNN running on the same (fixed) compression ratio. The energy overhead of the PTC-SNN is still near linearly scaled down by the compression ratio albeit that it is somewhat greater than that of the corresponding TC-SNN. And yet, the PT-SNN reduces the energy dissipation
and ATEL of the baseline by 6.59x and 16.53x, respectively when running at 16:1 compression ratio. 

\begin{table}
\caption{Performances of the reconfigurable PTC-SNN hardware accelerator on the TI46 Speech Corpus. }
\label{Performance_TI46_PTC}
\centering
\setlength{\tabcolsep}{2pt}
\begin{tabular}{llllll} 
\hline
\begin{tabular}[c]{@{}l@{}} Compres-\\sion ratio \end{tabular} & Accuracy & \begin{tabular}[c]{@{}l@{}} Power(W)\\~@50MHz \end{tabular} & \begin{tabular}[c]{@{}l@{}} Runti-\\me(s) \end{tabular} & \begin{tabular}[c]{@{}l@{}} Energy\\(J) \end{tabular} & \begin{tabular}[c]{@{}l@{}} Normali-\\zed ATEL \end{tabular}  \\ 
\hline
Baseline                                                       & 96.15\%  & 0.073                                                      & 1.991                                                   & 0.145                                                & 100\%                                                         \\
2:1                                                            & 96.15\%  & 0.151                                                      & 0.995                                                   & 0.130                                                & 57.64\%                                                       \\
3:1                                                            & 92.31\%  & 0.152                                                      & 0.664                                                   & 0.088                                                & 51.65\%                                                       \\
4:1                                                            & 92.31\%  & 0.155                                                      & 0.499                                                   & 0.067                                                & 29.87\%                                                       \\
8:1                                                            & 86.54\%  & 0.173                                                      & 0.248                                                   & 0.038                                                & 14.61\%                                                       \\
16:1                                                           & 80.77\%  & 0.194                                                      & 0.125                                                   & 0.022                                                & 6.05\%                                                        \\
\hline
\end{tabular}
\end{table}

\section{Conclusion}
We propose a general time compression technique and  two compression architectures, namely TC-SNN and PTC-SNN, to significantly boost the throughput and reduce energy dissipation of SNN accelerators. Our experimental results show that the proposed time compression architectures can support large time compression ratios of up to 16$\times$, delivering up to 15.93$\times$, 13.88$\times$, and 86.21$\times$ improvements in throughput, energy dissipation,  and a figure of merit (ATEL), respectively, and be realized  with modest additional hardware design overhead. 

\bibliographystyle{plain}

\end{document}